\documentclass{article}

\usepackage{graphicx}
\usepackage{pdfpages}
\usepackage{amsmath}
\usepackage{amssymb}
\usepackage{mathtools}
\usepackage{MnSymbol}
\usepackage{enumerate}
\usepackage{setspace}
\usepackage[hyphens]{url}
\usepackage{float}
\usepackage{caption}
\usepackage{subcaption}
\usepackage{multicol}
\usepackage{color}
\usepackage{listings}
\usepackage{csvsimple}
\usepackage{algorithm}
\usepackage{algorithmic}
\usepackage{verbatim}
\usepackage{mdframed}
\usepackage{changepage}
\usepackage[numbers]{natbib}

\usepackage[final]{neurips_2019}

\usepackage[utf8]{inputenc} 
\usepackage[T1]{fontenc}    
\usepackage{hyperref}       
\usepackage{url}            
\usepackage{booktabs}       
\usepackage{amsfonts}       
\usepackage{nicefrac}       
\usepackage{microtype}      

\title{Proposed Guidelines for the Responsible Use of\\ Explainable Machine Learning}

\author{Patrick Hall\\
H2O.ai\\
Washington, DC\\
\texttt{phall@h2o.ai} \And
Navdeep Gill\\
H2O.ai\\
Mountain View, CA\\
\texttt{navdeep@h2o.ai} \And
Nicholas Schmidt\\
BLDS, LLC\\
Philadelphia, PA\\
\texttt{nschmidt@bldsllc.com}
}

\newcommand{\twopartdef}[4]
{
	\left\{
		\begin{array}{ll}
			#1 & \mbox{if } #2 \\
			#3 & \mbox{if } #4
		\end{array}
	\right.
}

\begin{document}

\maketitle

\section{Introduction}

Explainable machine learning (ML) enables human learning from ML, human appeal of automated model decisions, regulatory compliance, and security audits of ML models.\footnote{\scriptsize{This text and associated software are not, and should not be construed as, legal advice or requirements for regulatory compliance.}}\textsuperscript{,}\footnote{\scriptsize{In the U.S., interpretable models, explanations, disparate impact testing, and the model documentation they enable may be required under the Civil Rights Acts of 1964 and 1991, the Americans with Disabilities Act, the Genetic Information Nondiscrimination Act, the Health Insurance Portability and Accountability Act, the Equal Credit Opportunity Act (ECOA), the Fair Credit Reporting Act (FCRA), the Fair Housing Act, Federal Reserve SR 11-7, and the European Union (E.U.) Greater Data Privacy Regulation (GDPR) Article 22 \cite{ff_interpretability}}.\label{fn:regs}}\textsuperscript{,}\footnote{\scriptsize{For various security applications, see: \href{https://www.oreilly.com/ideas/proposals-for-model-vulnerability-and-security}{``Proposals for Model Vulnerability and Security''}.}} Explainable ML (i.e. e\textit{x}plainable \textit{a}rtificial \textit{i}ntelligence or XAI) has been implemented in numerous open source and commercial packages and explainable ML is also an important, mandatory, or embedded aspect of commercial predictive modeling in industries like financial services.\footnote{\scriptsize{E.g. open source software listed here: \href{https://github.com/jphall663/awesome-machine-learning-interpretability}{``Awesome machine learning interpretability''}.}}\textsuperscript{,}\footnote{\scriptsize{E.g.,  Datarobot, H2O Driverless AI, SAS Visual Data Mining and Machine Learning, Zest AutoML.}}\textsuperscript{,}\footnote{\scriptsize{E.g., \href{https://ww2.amstat.org/meetings/jsm/2019/onlineprogram/AbstractDetails.cfm?abstractid=303053}{``Deep Insights into Explainability and Interpretability of Machine Learning Algorithms and Applications to Risk Management''}.}\label{fn:chen}} However, like many technologies, explainable ML can be misused, particularly as a faulty safeguard for harmful black-boxes, e.g. \textit{fairwashing} or \textit{scaffolding}, and for other malevolent purposes like stealing models and sensitive training data \cite{fair_washing}, \cite{please_stop}, \cite{membership_inference}, \cite{scaffolding}, \cite{model_stealing}. To promote best-practice discussions for this already in-flight technology, this short text presents internal definitions and a few examples in Section \ref{sec:intro} before covering the proposed guidelines in Subsections \ref{sec:trust} -- \ref{sec:white_box}. This text concludes in Section \ref{sec:conclusion} with a seemingly natural argument for the use of interpretable models and explanatory, debugging, and disparate impact testing methods in life- or mission-critical ML systems.

\section{Definitions and Examples} \label{sec:intro}

While the explainable ML community has apparently not yet adopted a clear taxonomy of concepts or a precise vocabulary, many authors have grappled with ideas related to interpretability and explanations. Some of these efforts include: ``A Survey of Methods for Explaining Black Box Models''  by \citet{guidotti2018survey}, ``The Mythos of Model Interpretability'' by \citet{lipton1}, \textit{\textbf{Interpretable Machine Learning}} by \citet{molnar}, ``Interpretable Machine Learning: Definitions,
Methods, and Applications'' by \citet{murdoch2019interpretable}, and ``Challenges for Transparency''  by \citet{weller2017challenges}. To decrease ambiguity herein, this section uses the review and survey corpus and practical examples to address the terms and phrases \textit{interpretable}, \textit{explanation}, \textit{explainable ML}, \textit{interpretable models}, \textit{model debugging techniques}, \textit{unwanted sociological bias}, and \textit{fairness techniques} before proposing guidelines.

\subsection{Interpretable and Explanation}

\citet{been_kim1} define interpretability in ML as, ``the ability to explain or to present in understandable terms to a human.'' Professor Sameer Singh of the University of California at Irvine (UCI), co-inventor of the seminal local interpretable model-agnostic explanation (LIME) technique, defines \textit{explanation} as a, ``collection of visual and/or interactive artifacts that provide a user with sufficient description of a model's behavior to accurately perform tasks like evaluation, trusting, predicting, or improving a model.''\footnote{\scriptsize{From: \href{https://github.com/jphall663/kdd_2019}{``Proposed Guidelines for the Responsible Use of Explainable Machine Learning''} (presentation only).}} And \citet{gilpin2018explaining} posit that a \textit{good explanation} occurs when modelers or consumers ``can no longer keep asking why'' in regards to some ML model behavior. These three thoughtful characterizations link explainability to interpretability, give clarity on explanation, and provide an abstract goal for any explainability task.

\subsection{Explainable ML and Interpretable Models }

Herein \textit{explainable ML} means mostly post-hoc analysis and techniques used to understand trained model mechanisms or predictions. Examples of common explainable ML techniques include:

\begin{itemize}
\item Local and global feature importance, e.g., Shapley and derivative-based feature attribution \cite{grad_attr} \cite{keinan2004fair}, \cite{shapley}, \cite{shapley1988shapley}, \cite{kononenko2010efficient}.
\item Local and global model-agnostic surrogate models, e.g., surrogate decision trees and LIME \cite{dt_surrogate2}, \cite{viper}, \cite{dt_surrogate1}, \cite{lime-sup}, \cite{lime}, \cite{wf_xnn}.
\item Local and global visualizations of model predictions, e.g., accumulated local effect (ALE) plots, 1- and 2-dimensional partial dependence plots, and individual conditional expectation (ICE) plots \cite{ale_plot}, \cite{esl}, \cite{ice_plots}.
\end{itemize}

Although difficult to quantify, credible research efforts into scientific measures of interpretability are underway \cite{friedler2019assessing}, \cite{molnar2019quantifying}, and the ability to measure degrees of interpretability implies it is not a binary, on-off quantity. Here, unconstrained, traditional black-box ML models, such as multilayer perceptron (MLP) neural networks and gradient boosting machines (GBMs), are said to be difficult to interpret, potentially unsafe for use in life- or mission-critical applications, but not necessarily completely unexplainable. In this text, \textit{interpretable models} (i.e., white-box models) will include linear models, decision trees, rule-based models, constrained or Bayesian variants of traditional black-box ML models, or novel types of models designed to be directly interpretable. Examples of newer, highly interpretable ML modeling techniques include explainable neural networks (XNNs), explainable boosting machines (EBMs, GA2Ms), monotonically constrained GBMs, scalable Bayesian rule lists, or super-sparse linear integer models (SLIMs), \cite{ga2m}, \cite{slim}, \cite{wf_xnn}, \cite{sbrl}.\footnote{\scriptsize{EBM, as implemented in the \href{https://github.com/microsoft/interpret}{Microsoft \texttt{interpret} package}.}}\textsuperscript{,}\footnote{\scriptsize{Monotonic GBM, as implemented in \href{https://xgboost.readthedocs.io/en/latest/tutorials/monotonic.html}{\texttt{XGBoost}} or \href{https://github.com/h2oai/h2o-3/blob/master/h2o-py/demos/H2O_tutorial_gbm_monotonicity.ipynb}{\texttt{h2o}}.}}\textsuperscript{,}\footnote{\scriptsize{And similar methods, e.g.: \url{https://users.cs.duke.edu/~cynthia/papers.html}}.}

\subsection{Model Debugging Techniques}

Herein \textit{model debugging techniques} test ML models to increase trust in mechanisms and predictions. Debugging techniques include model assertions, security audits, variants of sensitivity (i.e., \textit{what-if?}) analysis, variants of residual analysis and residual explanation, and unit tests to verify the accuracy or security of ML models \cite{modeltracker}, \cite{kangdebugging}.\footnote{\scriptsize{And similar methods, e.g.: \url{https://debug-ml-iclr2019.github.io/}.}} Model debugging should also include remediating any discovered errors or vulnerabilities.

\subsection{Unwanted Sociological Bias and Fairness Techniques}

In this text, \textit{unwanted sociological bias} encompasses several forms of discrimination that may manifest in ML, including overt discrimination, disparate treatment, and disparate impact (DI), i.e., unintentional discrimination. DI may be caused by model misspecification, inaccurate or incomplete data, or data that has differing correlations or dependencies among demographic groups of individuals, driving differences in favorable model outcomes. A model is said to be biased here if, (1) group membership is not independent of the likelihood of a favorable outcome, or (2) under certain circumstances, membership in a \textit{subset} of a group is not independent of the likelihood of a favorable outcome (i.e., \textit{local} bias). Underlying discrimination that causes bias may or may not be illegal, depending on how it arises and applicable discrimination laws. Herein \textit{fairness techniques} are used to diagnose and remediate unwanted sociological bias in ML models. Diagnosis approaches include DI testing and other tests for bias \cite{feldman2015certifying}. Remediation methods tend to involve model selection by minimization of bias, preprocessing training data, e.g., reweighing \cite{kamiran2012data}, training unbiased models, e.g., adversarial de-biasing \cite{zhang2018mitigating}, or post-processing model predictions, e.g., by equalized odds \cite{hardt2016equality}.\footnote{\scriptsize{And similar methods, e.g.: \url{http://www.fatml.org/resources/relevant-scholarship}.}}

\section{Proposed Guidelines for the Responsible Use of Explainable ML}

Four guidelines are proposed and discussed in Subsections \ref{sec:trust} -- \ref{sec:white_box} to assist practitioners in avoiding unintentional misuse or identifying intentional abuse of explainable ML. The guidelines are:

\begin{enumerate}

\item Use explanations to enable understanding.
\item Learn how explainable ML is used for nefarious purposes.
\item Augment surrogate models with direct explanations.
\item Use highly interpretable mechanisms for life- or mission-critical ML.

\end{enumerate}

\noindent Important corollaries to the guidelines are also highlighted and simple, reproducible software examples accompany the guidelines to avoid hypothetical reasoning whenever possible.

\subsection{Guideline: Use Explanations to Enable Understanding} \label{sec:trust}

\begin{figure}[htb!]
	\begin{subfigure}{.4\textwidth}
  		\includegraphics[height=1.17\linewidth, width=1.05\linewidth]{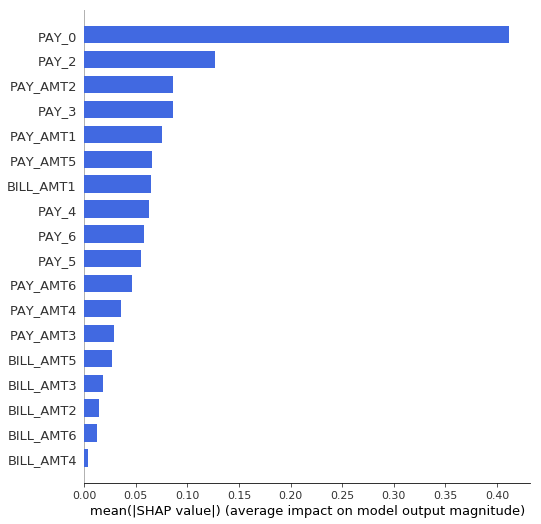}
  		\caption{Consistent global Shapley feature importance values for $g_{\text{GBM}}$.}
  		\label{fig:global_shap}
	\end{subfigure}\hspace{10pt}
	\begin{subfigure}{.5\textwidth}
		\includegraphics[width=1.28\linewidth]{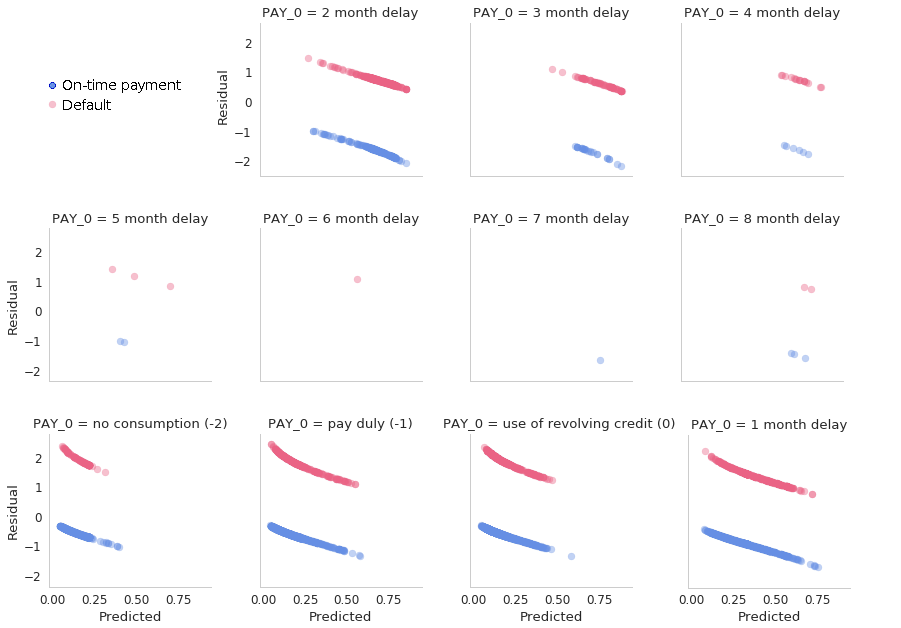}
  		\caption{$g_{\text{GBM}}$ deviance residuals and predictions by \texttt{PAY\_0}.}
  		\label{fig:resid}
	\end{subfigure}
	\caption{An unconstrained GBM probability of default model, $g_{\text{GBM}}$, generally over-emphasizes the importance of the input feature \texttt{PAY\_0}, a customer's most recent repayment status, in the UCI credit card data. $g_{\text{GBM}}$ often produces large positive residuals when \texttt{PAY\_0} indicates on-time payments (\texttt{PAY\_0} $\leq$ 1) and large negative residuals when \texttt{PAY\_0} indicates late payments (\texttt{PAY\_0} $>$ 1). Combining explanatory and debugging techniques shows that $g_{\text{GBM}}$ is explainable, but probably not trustworthy.}
	\label{fig:global_shap_resid}
\end{figure}

Explanations are often discussed in the context of trust (e.g., \citet{lime}), but explanations alone are not sufficient for trust in ML models. Explanation, as a general concept, is related more directly to understanding and transparency than to trust.\footnote{\scriptsize{The \href{https://www.merriam-webster.com/dictionary/explain}{Merriam-Webster definition} of \textit{explain}, accessed Sept. 8\textsuperscript{th} 2019, does not mention \textit{trust}.}} Simply put, one can understand and explain a model without trusting it. One can also trust a model and not be able to understand or explain it. Consider the following example scenarios.

\begin{itemize}

\item \textbf{Explanation and understanding without trust}: In Figure \ref{fig:global_shap_resid}, global Shapley explanations and residual analysis identify a pathology in an unconstrained GBM model, $g_{\text{GBM}}$, trained on the UCI credit card dataset \cite{uci}.\footnote{\scriptsize{Code to replicate Figure \ref{fig:global_shap_resid}: \url{https://bit.ly/2m58Lxl}.}} $g_{\text{GBM}}$ over emphasizes the input feature \texttt{PAY\_0}, or a customer's most recent repayment status. Due to over-emphasis of \texttt{PAY\_0}, $g_{\text{GBM}}$ is often unable to predict on-time payment if recent payments are delayed (\texttt{PAY\_0} $>$ 1), causing large negative residuals. $g_{\text{GBM}}$ is also often unable to predict default if recent payments are made on-time (\texttt{PAY\_0} $\leq$ 1), causing large positive residuals. In this example scenario, $g_{\text{GBM}}$ is explainable, but likely untrustworthy.

\item \textbf{Trust without explanation and understanding}: Years before reliable explanation techniques were widely acknowledged and available, black-box predictive models, such as autoencoder and MLP neural networks, were used for fraud detection in the financial services industry \cite{gopinathan1998fraud}. When these models performed well, they were trusted.\footnote{\scriptsize{E.g., \href{https://www.sas.com/en_ph/customers/hsbc.html}{``Reduce Losses from Fraudulent Transactions''}.}}\textsuperscript{,}\footnote{\scriptsize{E.g., \href{https://www.kdnuggets.com/2011/03/sas-patent-fraud-detection.html}{``SAS Secures Technology Patent for Better Fraud Detection Performance''}.}} However, they were not explainable or well-understood by contemporary standards.

\end{itemize}
Explanations typically increase trust in models as a side-effect when they are acceptable to human users by various criteria. As illustrated in Figure \ref{fig:hc_ml}, in an ideal scenario, explanation techniques should be used to directly increase understanding in ML models, while debugging and DI testing methods should be used to directly promote trust.

\subsection{Guideline: Learn How Explainable ML is Used for Nefarious Purposes}

When used disingenuously, explainable ML methods can provide cover for misused or intentionally abusive black-boxes \cite{fair_washing}, \cite{please_stop}, \cite{scaffolding}. Explainable ML methods can also enable hacking or stealing of models or data through public prediction APIs or other endpoints \cite{membership_inference}, \cite{model_stealing}. Moreover, explainable ML methods are likely to be used for other nefarious purposes in the future and may be used for unknown destructive purposes now. Responsible practitioners need to understand the malevolent side of this technology to better detect and correct misuse and abuse.

\subsubsection{Corollary: Use Explainable ML for Security Audits} Use explainable ML techniques to test ML systems for vulnerabilities to model stealing, inversion, and membership inference attacks.

\subsubsection{Corollary: Explainable ML Can be Used to Crack Nefarious Black-boxes} Used as white-hat hacking tools, explainable ML can help draw attention to accuracy or unwanted sociological bias problems in proprietary black-boxes. See \citet{angwin16} for evidence that cracking proprietary black-box models for oversight purposes is possible.\footnote{\scriptsize{This text makes no claim on the quality of the analysis in Angwin et al. (2016), which has been criticized \cite{flores2016false}. This now infamous analysis is presented only as evidence that motivated activists can crack proprietary black-boxes using surrogate models and other explanatory techniques. Moreover, such analyses would likely improve with established best-practices for explainable ML.}}

\subsubsection{Corollary: Explainable ML is a Privacy Vulnerability} Recent research shows that providing explanations along with predictions eases attacks that can compromise sensitive training data \cite{shokri2019privacy}.

\subsection{Guideline: Augment Surrogate Models with Direct Explanations}

Models of models, or surrogate models, can be helpful explanatory tools, but they are usually approximate, low-fidelity explainers. Aside from (1) a global or local summary of a complex model provided by a surrogate model can be helpful sometimes and (2) much work in explainable ML has been directed toward improving the fidelity and usefulness of surrogate models \cite{dt_surrogate2}, \cite{viper}, \cite{dt_surrogate1}, \cite{lime-sup}, \cite{wf_xnn}, \textit{many explainable ML techniques have nothing to do with surrogate models}. One of the most exciting breakthroughs for supervised learning problems in explainable ML is the application of a coalitional game theory concept, Shapley values, to compute feature attributions which are consistent globally and accurate locally using the trained model itself \cite{shapley}, \cite{kononenko2010efficient}. An extension of this idea, called Tree SHAP, has already been implemented for popular tree ensemble methods \cite{tree_shap}.

\begin{figure}[htb!]
	\begin{subfigure}{.55\textwidth}
		\includegraphics[height=0.6\linewidth, width=.95\linewidth]{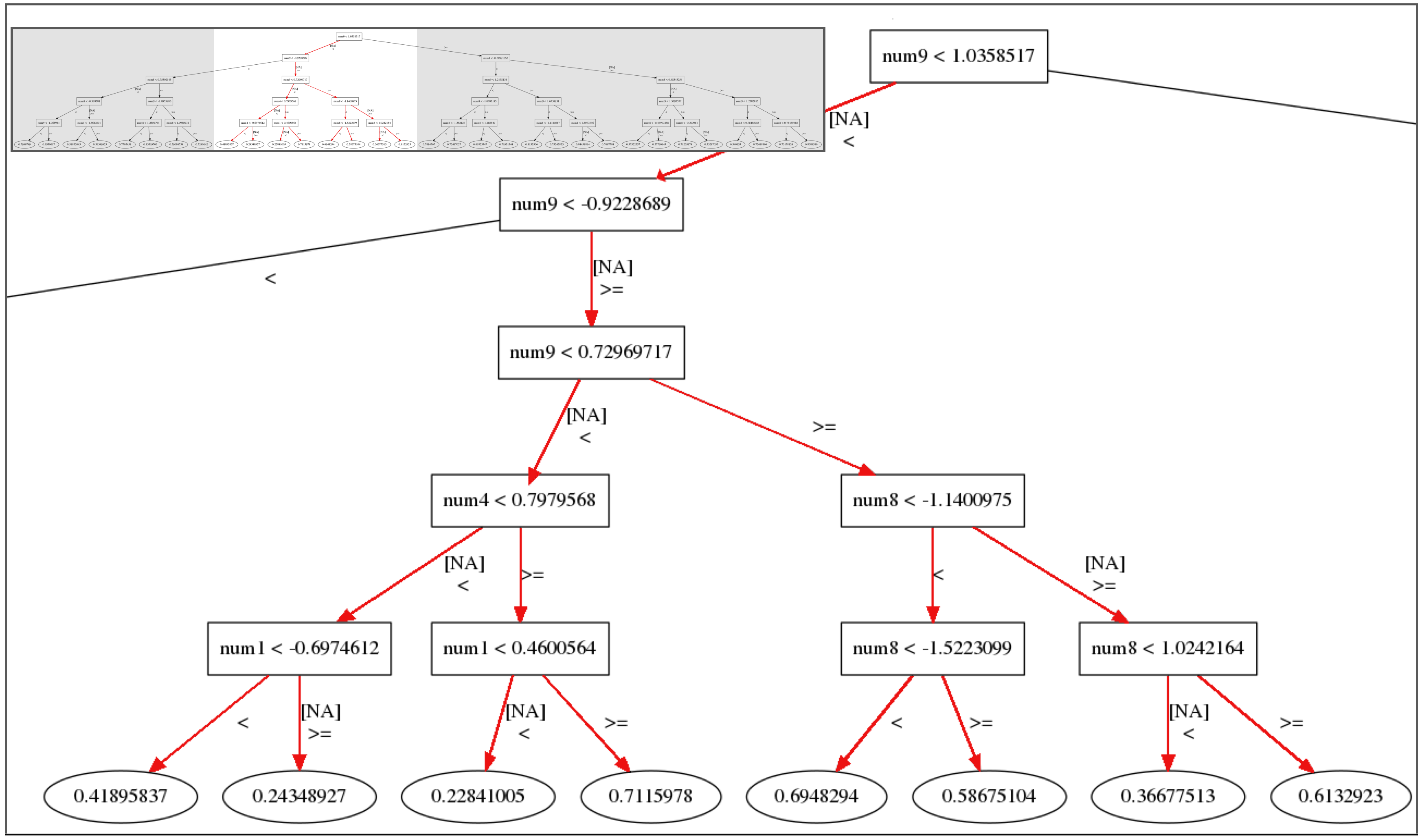}
  		\caption{Na\"ive $h_{\text{tree}}$, \textit{a surrogate model}, forms an approximate overall flowchart for the explained model, $g_{\text{GBM}}$.}
  		\label{fig:dt_surrogate}
	\end{subfigure}\hspace{5pt}
	\begin{subfigure}{.45\textwidth}
  		\includegraphics[height=.52\linewidth, width=1.02\linewidth]{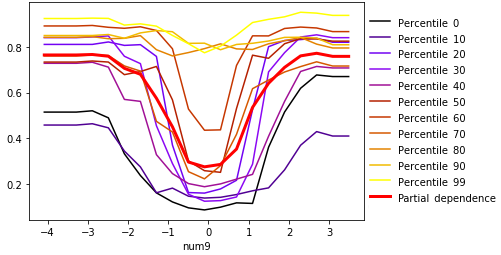}
  		\caption{Partial dependence and ICE curves generated \textit{directly from the explained model}, $g_{\text{GBM}}$.}
  		\label{fig:pdp_ice}
	\end{subfigure}
	\caption{$h_{\text{tree}}$ displays known interactions in $f = X_{\text{num}1} * X_{\text{num}4} + |X_{\text{num}8}| * X_{\text{num}9}^2$ for $\sim -1 < X_{\text{num9}} <  \sim 1$. Modeling of the known interactions in $f$ by $g_{\text{GBM}}$ is also highlighted by the divergence of partial dependence and ICE curves for $\sim -1 < X_{\text{num9}} <  \sim 1$. Explanations from a surrogate model have augmented and confirmed findings from a direct model visualization technique.}
	\label{fig:pdp_ice_dt_surrogate}
\end{figure}

There are many other explainable ML methods that operate on trained models directly such as partial dependence, ALE, and ICE plots \cite{ale_plot}, \cite{esl}, \cite{ice_plots}. Surrogate models and explanatory techniques that operate directly on trained models can also be combined, for instance by using partial dependence, ICE, and surrogate decision trees to investigate and confirm modeled interactions \cite{art_and_sci}. In Figure \ref{fig:pdp_ice_dt_surrogate}, an unconstrained GBM, $g_{\text{GBM}}$, models a known signal generating function $f$:

\begin{equation}
\label{eq:f}
f(\mathbf{X}) = \twopartdef {1} {X_{\text{num}1} * X_{\text{num}4} + |X_{\text{num}8}| * X_{\text{num}9}^2 + e \geq 0.42} {0} {X_{\text{num}1} * X_{\text{num}4} + |X_{\text{num}8}| * X_{\text{num}9}^2 + e < 0.42}
\end{equation}

\noindent where $e$ signifies the injection of random noise in the form of label switching for roughly 15\% of the training and validation observations.\footnote{\scriptsize{Code to replicate Figure \ref{fig:pdp_ice_dt_surrogate}: \url{https://bit.ly/2kSuAQD}.}} $g_{\text{GBM}}$ is then trained such that $g_{\text{GBM}}(\mathbf{X}) \approx f(\mathbf{X})$ in training and validation data. $h_{\text{tree}}$, displayed in Figure \ref{fig:dt_surrogate}, is extracted such that $h_{\text{tree}}(\mathbf{X}) \approx g_{\text{GBM}}(\mathbf{X}) \approx f(\mathbf{X})$ in validation data. Partial dependence and ICE plots are generated directly for $g_{\text{GBM}}$ in the same validation data and overlaid in Figure \ref{fig:pdp_ice}. The parent-child node relationships displayed in $h_{\text{tree}}$ for $\sim -1 < X_{\text{num9}} < \sim 1$ in \ref{fig:dt_surrogate} and the divergence of ICE and partial dependence curves in \ref{fig:pdp_ice} for $\sim -1 < X_{\text{num9}} < \sim 1$ help confirm and explain how $g_{\text{GBM}}$ learned the interactions in $f$. As in Figure \ref{fig:global_shap_resid}, combining different approaches provided additional, beneficial information about a ML model.

\subsubsection{Corollary: Augment LIME with Direct Explanations} LIME is important, imperfect (like every other ML technique), and vulnerable to adversarial manipulation \cite{scaffolding}. LIME, in its most popular implementation, uses local linear surrogate models fit to perturbed, locally weighted samples to explain regions of machine-learned decision boundaries or response functions \cite{lime}. Like other surrogate models, LIME can be combined with model-specific methods for validation and to yield deeper insights. Consider that Tree SHAP can provide locally accurate and consistent point estimates for local feature importance as in \ref{fig:shap} below. LIME can then provide approximate information about modeled local linear trends around the same point. Table \ref{tab:lime} contains LIME $h_{\text{GLM}}$ coefficients for a local region of a validation set sampled from the UCI credit card data defined by \texttt{PAY\_0 > 1}, or customers with a fairly high risk of default due to late most recent payments.\footnote{\scriptsize{Code to replicate Table \ref{tab:lime}: \url{https://bit.ly/2miCPpo}.}} $h_{\text{GLM}}$ models the predictions of a simple interpretable decision tree model, $g_{\text{tree}}$, displayed in \ref{fig:dt}. $h_{\text{GLM}}$ coefficients show linear trends between features in the sampled set $\mathbf{X}_{\text{PAY\_0} > 1}$ and $g_{\text{tree}}(\mathbf{X}_{\text{PAY\_0}> 1})$. Because $h_{GLM}$ is relatively well-fit (0.73 $R^2$) and has a logical intercept (0.77), it can be used along with Shapley values to reason about the modeled average behavior for risky customers, to differentiate the behavior of any one specific risky customer from their peers under the model, or to validate LIME results.

\begin{table}
  	\caption{Coefficients for local linear model, $h_{\text{GLM}}$, with an intercept of 0.77 and an $R^2$ of 0.73. $h_{\text{GLM}}$ is trained on a segment of the UCI credit card dataset containing higher-risk customers with late most recent repayment statuses, $\mathbf{X}_{PAY \_ 0 > 1}$, and the predictions of a decision tree, $g_{\text{tree}}(\mathbf{X}_{\text{PAY\_0} > 1})$.\\}
  	\label{tab:lime}
  	\centering
  	\begin{tabular}{ll}
    	\toprule
    	$h_{\text{GLM}}$ Feature & $h_{\text{GLM}}$ Coefficient \\
    	\midrule
		\texttt{PAY\_0 == 4} & $0.0009$ \\
		\texttt{PAY\_2 == 3} & $0.0065$ \\
		\texttt{PAY\_5 == 2} & $-0.0006$ \\
		\texttt{PAY\_6 == 2} & $0.0036$ \\
		\texttt{BILL\_AMT1} & $3.4339\mathrm{e}{-08}$ \\
		\texttt{PAY\_AMT1} & $4.8062\mathrm{e}{-07}$ \\
		\texttt{PAY\_AMT3} & $-5.867\mathrm{e}{-07}$ \\
    	\bottomrule
  \end{tabular}
\end{table}

\subsection{Guideline: Use Highly Interpretable Mechanisms for Mission- or Life-Critical ML} \label{sec:white_box}

Given the known difficulties with explaining black-boxes \cite{please_stop}, the existence of unwanted social bias in data and ML models \cite{fairmlbook}, the security vulnerabilities of ML (e.g., \citet{membership_inference}, \citet{model_stealing}), and the potentially surprising behavior of black-boxes (e.g., \citet{easily_fooled}, \citet{intriguing_properties}), it appears prudent today to use highly transparent ML mechanisms for  applications that make life-altering or high-value decisions. Interpretability, as enabled by interpretable models and post-hoc explanations (see Corollary \ref{cor:int_ex}), may be mandated by regulation for some life- or mission-critical applications, but interpretability is also recommended for any ML application in which inevitable wrong decisions should be appealable. This subsection discusses a few details and examples regarding regulated ML applications and appeal, and also advocates for trust-enhancing DI testing (see Corollaries \ref{cor:ex_di} -- \ref{cor:di_con}) in high-stakes or human-centered applications.

Interpretable ML mechanisms are required under numerous regulatory statutes in the U.S., and explainable ML tools like LIME and other surrogate models, partial dependence plots, and global and local feature importance are already used to document, understand, and validate some predictive models in the financial services industry \cite{lime-sup}, \cite{wf_xnn}.\textsuperscript{\ref{fn:regs}, \ref{fn:chen}} Moreover, adverse action notices are mandated under the Equal Credit Opportunity Act (ECOA) and the Fair Credit Reporting Act (FCRA) for many credit lending, employment, and insurance decisions in the U.S.\footnote{\scriptsize{See: \href{https://consumercomplianceoutlook.org/2013/second-quarter/adverse-action-notice-requirements-under-ecoa-fcra/}{``Adverse Action Notice Requirements Under the ECOA and the FCRA''}.}} If ML is used for such decisions, it must be explained in terms of adverse action notices.\footnote{\scriptsize{This is apparently already happening: \href{https://www.prnewswire.com/news-releases/new-patent-pending-technology-from-equifax-enables-configurable-ai-models-300701153.html}{``New Patent-Pending Technology from Equifax Enables Configurable AI Models''}.}} Shapley values, and other local feature importance approaches, provide a convenient methodology to rank the direct contribution of input features to final model decisions and potentially generate customer-specific adverse action notices.

Aside from regulatory mandates, interpretable models and explanations enable logical appeal processes for automated decisions made by ML models. Consider being negatively impacted by an erroneous black-box model decision, say for instance being mistakenly denied a loan or parole. How would you argue your case for appeal without knowing how model decisions were made? According to the New York Times, a man named Glenn Rodr\'iguez found himself in this unfortunate position in a penitentiary in Upstate New York in 2016.\footnote{\scriptsize{\href{https://www.nytimes.com/2017/06/13/opinion/how-computers-are-harming-criminal-justice.html}{See: ``When a Computer Program Keeps You in Jail''}.}}

\subsubsection{Corollary: Use Interpretable Models Along with Explanation Techniques} \label{cor:int_ex}

\begin{figure}[ht!]
	\begin{subfigure}{.6\textwidth}
		\includegraphics[height=.45\linewidth, width=1.15\linewidth]{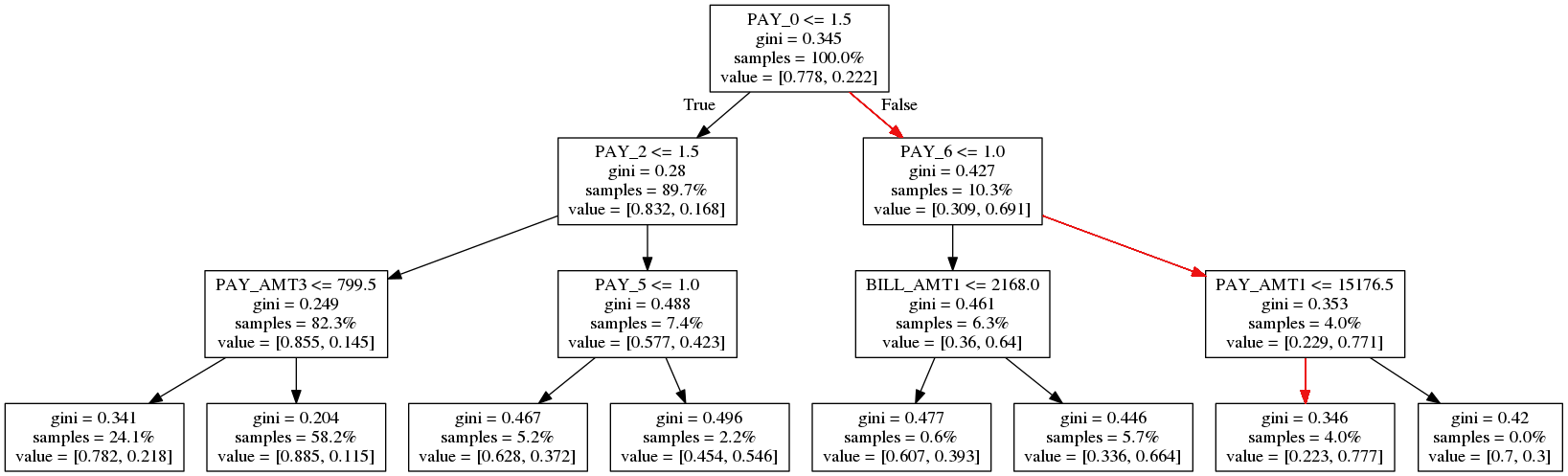}
  		\caption{Simple decision tree, $g_{\text{tree}}$, trained on the UCI credit card data to predict default with validation AUC of 0.74. The decision policy for a high-risk individual is highlighted in red.}
  		\label{fig:dt}
	\end{subfigure}\hspace{50pt}
	\begin{subfigure}{.4\textwidth}
		\vspace{22pt}
  		\includegraphics[height=.5\linewidth, width=.8\linewidth]{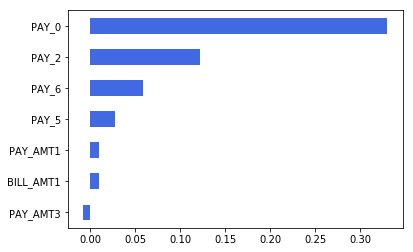}
  		\vspace{5pt}
  		\caption{Locally-accurate Shapley contributions\\ for the highlighted individual's probability\\ of default.}
  		\label{fig:shap}
	\end{subfigure}
	\caption{The decision-policy for a high-risk customer is highlighted in \ref{fig:dt} and the locally-accurate Shapley contributions for this same individual's predicted probability are displayed in \ref{fig:shap}. Due to their consistency properties, Shapley values highlight the local importance of features not on the decision path in this particular encoding, i.e., $g_{\text{tree}}$, of the unknown signal-generating function, which is likely helpful for the generation of consistent adverse action notices across similar data and models.}
	\label{fig:dt_shap}
\end{figure}

Some well-known publications have focused either on interpretable models (e.g., \citet{slim}, \citet{sbrl}) or on post-hoc explanations (e.g., \citet{shapley}, \citet{lime}), but the two can be used together in the context of a holistic ML workflow, illustrated in Figure \ref{fig:hc_ml}. Consider the seemingly useful example case of augmenting globally interpretable models with local post-hoc explanations. A practitioner could train a single decision tree, a globally interpretable model, then apply local explanations in the form of Shapley feature importance as in Figure \ref{fig:dt_shap}.\footnote{\scriptsize{Code to replicate Figure \ref{fig:dt_shap}: \url{https://bit.ly/2miCPpo}.}} This enables practitioners to see accurate numeric feature contributions for each prediction and the entire directed graph of the decision tree. Even for interpretable models, such as linear models and decision trees, Shapley values present accuracy and consistency advantages over standard feature attribution methods \cite{lipovetsky2001analysis}, \cite{shapley}, \cite{tree_shap}. Shapley values also enable the consistent ranking of input features for each model decision, which is likely helpful for FCRA and ECOA compliance. Another twist on the idea of combining explainable ML methods and interpretable models is described by \citet{gosiewska2019safe} in ``Surrogate Assisted Feature Extraction for Machine Learning (SAFE ML)''. In the SAFE ML approach, features learned by more complex models are extracted and used in an explainable fashion to increase the accuracy of more interpretable models. Aren't either of these augmented processes more desirable than either an interpretable model or post-hoc explanations by themselves?

\subsubsection{Corollary: Use Explanations Along with DI Testing} \label{cor:ex_di}

Like interpretable models, fairness methods are often presented in different articles than post-hoc explanatory methods. However, in banks for example, using post-hoc explanatory tools along with DI testing is often necessary to comply with model documentation guidance and with fair lending regulations.\footnote{\scriptsize{See: \href{https://www.ffiec.gov/pdf/fairlend.pdf}{``Interagency Fair Lending Examination Procedures''}.}}\textsuperscript{,}\footnote{\scriptsize{See: \href{https://files.consumerfinance.gov/f/documents/201510_cfpb_ecoa-narrative-and-procedures.pdf}{``CFPB Consumer Laws and Regulations: ECOA''}.}} To clarify, explanatory techniques should \textit{not} replace DI testing for bias detection purposes, but in general, explanations increase transparency and understanding of model mechanisms and predictions, while DI auditing and remediation increases trust that model predictions are as fair as possible. As in previous sections, trust and understanding are different but complimentary goals achieved by combining multiple approaches.

\begin{table}
  	\caption{Basic group disparity metrics for females under monotonically constrained GBM model, $g_{\text{mono}}$, trained on the UCI credit card dataset.\\}
  	\label{tab:dia}
  	\centering
  	\begin{tabular}{lll}
    	\toprule
    	& Adverse Impact Ratio & False Omissions Rate Disparity \\
    	\midrule
		Female & 0.8869 & \textcolor{red}{0.7866} \\
    	\bottomrule
  \end{tabular}
\end{table}

Table \ref{tab:dia} displays basic group disparity metrics for a monotonic GBM model, $g_{\text{mono}}$, trained on the UCI credit card data.\footnote{\scriptsize{Code to replicate Table \ref{tab:dia}:  \url{https://bit.ly/2lZUlyN}.} \label{fn:dia}} In this example, $g_{\text{mono}}$ displays group parity for adverse impact with \texttt{male} as the reference level according to the four-fifths rule, but also presents unwanted false omissions rate bias against females, indicating that males may be receiving too much credit they cannot repay, potentially preventing females from receiving that credit.\footnote{\scriptsize{The four-fifths rule was delineated by the Equal Employment Opportunity Commission (EEOC) as a measure of DI that would be of concern to regulators. This threshold has been associated with a specific measure of DI, the adverse impact ratio (AIR). While it may be applied to other measures of fairness, as in Table 2, this is often irrelevant in real-world compliance and litigation settings in the U.S.}} This disparity can be remedied by gently increasing the decision cutoff for $g_{\text{mono}}$, and Shapley values can also explain each $g_{\text{mono}}$ prediction.\textsuperscript{\ref{fn:dia}} Beyond explaining predictions, Explainable ML can assist in the difficult problem of determining input features within wide training sets that drive DI. For example, weighted average Shapley values can be analyzed by demographic segment, highlighting the features that have the largest deleterious impact on the protected segment. Explainable ML techniques, especially when paired with clustering, can also be useful for isolating instances of local bias.

\subsubsection{Corollary: Explanation is Not a Frontline Fairness Tool} Demographic attributes cannot currently be used in predictive models for high-stakes and commercially viable uses of explainable ML in credit lending, insurance, and employment in the U.S. that fall under FCRA, ECOA, or other applicable regulations. Thus their contribution to models cannot be assessed using accurate, direct explainable ML techniques. Even when demographic attributes can be used in models, it has been shown that explanations may not detect unwanted sociological bias \cite{fair_washing}. Given these drawbacks, it is recommended that fairness techniques are used to test for and remediate bias, and explanations are used to understand bias when appropriate (see Corollary \ref{cor:ex_di}).

\subsubsection{Corollary: Use DI Testing Along with Constrained Models} \label{cor:di_con}

Unconstrained ML models can treat similar individuals differently due to small differences in input data values, causing local bias that is not detectable with standard DI testing methods that measure group fairness \cite{dwork2012fairness}. To mitigate local bias when using ML, and to ensure standard bias or DI testing methods are most effective, pair such testing with constrained models.

\section{Conclusion: a Holistic Approach for Life- or Mission-Critical ML} \label{sec:conclusion}

\begin{figure}[htb!]
	\begin{center}
		\includegraphics[scale=0.1]{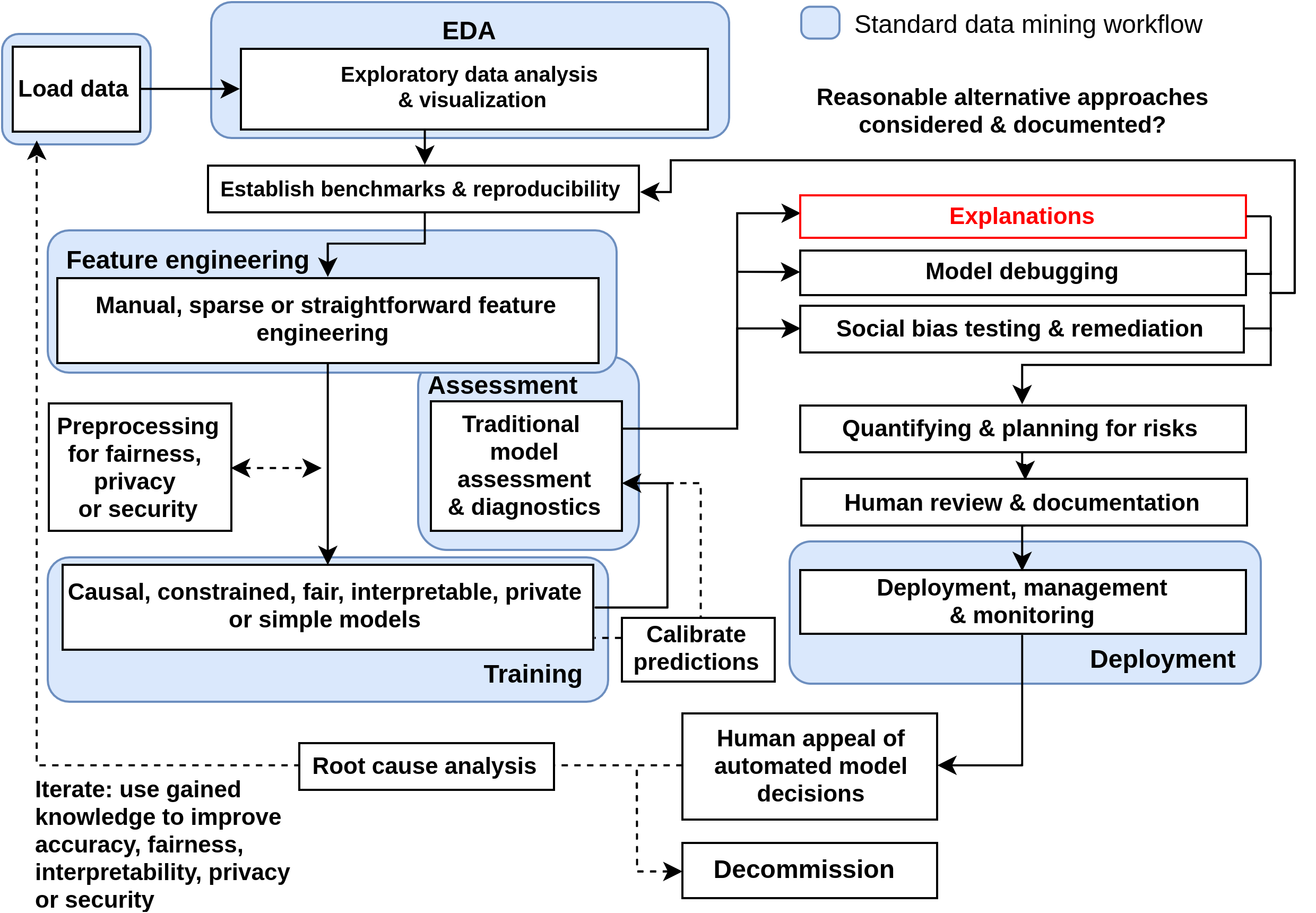}
		\caption{A diagram of a proposed holistic ML workflow in which explanations (highlighted in red) are used along with interpretable models, DI analysis and remediation techniques, and other review and appeal mechanisms to create an understandable and trustworthy ML system.}
		\label{fig:hc_ml}
	\end{center}
\end{figure}

ML systems are used today to make life-altering decisions about employment, bail, parole, and lending,\footnote{\scriptsize{See: \href{https://debug-ml-iclr2019.github.io/}{``Debugging Machine Learning Models''}.}} and the scope of decisions delegated to ML systems seems likely to expand in the future. Many researchers and practitioners are tackling DI, inaccuracy, privacy violations, and security vulnerabilities with a number of brilliant, but sometimes siloed, approaches. By proposing some straightforward explainable ML guidelines, this short text also gives examples of combining innovations from several sub-disciplines of ML research to train understandable and trustworthy predictive modeling systems. As illustrated in Figure \ref{fig:hc_ml}, these innovations can be used together, and this combination may be better-suited than conventional ML methods for use in business- and life-critical applications.

\clearpage
\section*{Acknowledgements}

The authors thank Przemyslaw Biecek, Pramit Choudhary, Benjamin Cox, Lingyao Meng, Christoph Molnar, Sameer Singh, and Bryce Stephens for their helpful input and insights.

\section*{References}
\small
\bibliographystyle{plainnat}
\bibliography{responsible_xai}

\end{document}